\documentclass[11pt,a4paper]{article}
\usepackage[hyperref]{acl2020}
\usepackage{times}
\usepackage{latexsym}

\usepackage{arydshln}
\usepackage{booktabs}
\usepackage{graphicx}
\usepackage{float}
\usepackage{color}
\usepackage{amsmath,amsfonts,amssymb,bbm,epsfig,bm}
\usepackage[ruled,noresetcount,linesnumbered]{algorithm2e}
\usepackage{listings}
\usepackage{subfig}
\usepackage{url}
\usepackage{spverbatim}
\usepackage{multirow}
\usepackage{pbox}
\usepackage{tikz}
\usepackage{pifont}
\usepackage{xspace}
\usepackage{subfig}
\usepackage{physics}
\usepackage[nameinlink,capitalize]{cleveref}
\usepackage{scalerel}
\usepackage[font=small]{caption}
\usepackage{titlesec}
\usepackage{todonotes}
\usepackage[normalem]{ulem}

\usepackage{microtype}

\usetikzlibrary{tikzmark}

\def\model/{\textsc{TaBert}}
\def\modelbase/{$\textsc{TaBert}_\textrm{\tt Base}$}
\def\modellarge/{$\textsc{TaBert}_\textrm{\tt Large}$}
\def\bert/{$\textsc{Bert}$}
\def\bertbase/{$\textsc{Bert}_\textrm{\tt Base}$}
\def\bertlarge/{$\textsc{Bert}_\textrm{\tt Large}$}
\newcommand{\CS}[1]{\ensuremath{\mathrm{(K=#1)}}}

\definecolor{darkgreen}{RGB}{43,163,39}
\definecolor{amaranth}{rgb}{0.9, 0.17, 0.31}

\renewcommand{\tt}[1]{\fontfamily{cmtt}\selectfont #1}

\newcommand{\sy}[1]{{\color{purple} [Scott: {#1}]}}
\newcommand{\pcyin}[1]{{\color{blue} [Pengcheng: {#1}]}}
\newcommand{\gn}[1]{{\color{darkgreen} [Graham: {#1}]}}

\renewcommand{\sy}[1]{}
\renewcommand{\pcyin}[1]{}
\renewcommand{\gn}[1]{}

\newcommand{\eg}{\hbox{\emph{e.g.}}\xspace}
\newcommand{\ie}{\hbox{\emph{i.e.}}\xspace}
\newcommand\mask[1]{\ensuremath{\widetilde{#1}}}

\newcommand\tbl{\ensuremath{{T}}}
\newcommand\tblvec{\ensuremath{\mathbf{T}}}
\newcommand\mr{\ensuremath{\bm{z}}}
\newcommand\utt{\ensuremath{\bm{u}}}
\newcommand\xvec{\ensuremath{\mathbf{x}}}

\newcommand\xseq{\ensuremath{\bm{x}}}
\newcommand\cell{\ensuremath{s}}
\newcommand\cellvec{\ensuremath{\mathbf{s}}}
\newcommand\col{\ensuremath{c}}
\newcommand\colvec{\ensuremath{\mathbf{c}}}
\newcommand\row{\ensuremath{R}}

\newcommand\std[1]{\ensuremath{\scriptstyle \pm #1}}

\def\wtq/{\textsc{WikiTableQuestions}}
\def\spider/{\textsc{Spider}}

\aclfinalcopy 
\def\aclpaperid{2876} 

\title{\model/: Pretraining for Joint Understanding of \\Textual and Tabular Data}

\author{
	Pengcheng Yin\thanks{~~Work done while at Facebook AI Research.} \quad Graham Neubig \\ 
	Carnegie Mellon University \\
	\tt{\{pcyin,gneubig\}@cs.cmu.edu}
	\And Wen-tau Yih \quad Sebastian Riedel \\
	Facebook AI Research \\
	\tt{\{scottyih,sriedel\}@fb.com}
}

\date{}

\begin{document}
\maketitle
\begin{abstract}
Recent years have witnessed the burgeoning of pretrained language models (LMs) for text-based natural language (NL) understanding tasks.
Such models are typically trained on free-form NL text, hence may not be suitable for 
tasks like semantic parsing over structured data, which require reasoning over both free-form NL questions and structured tabular data (\eg, database tables).
In this paper we present \model/, a pretrained LM that jointly learns representations for NL sentences and \mbox{(semi-)structured} tables.
\model/ is trained on a large corpus of 26 million tables and their English  contexts.
In experiments, neural semantic parsers using \model/ as feature representation layers achieve new best results on the challenging weakly-supervised semantic parsing benchmark \wtq/, while performing competitively on the text-to-SQL dataset \spider/.\footnote{Code available at \href{http://fburl.com/TaBERT}{\tt http://fburl.com/TaBERT}}

\end{abstract}

\section{Introduction}


Recent years have witnessed a rapid advance in the ability to understand and answer questions about free-form natural language (NL) text~\cite{Rajpurkar2016SQuAD10},
largely due to large-scale, pretrained language models (LMs) like BERT~\cite{Devlin2019BERTPO}.
These models allow us to capture the syntax and semantics of text via representations learned in an unsupervised manner, before fine-tuning the model to downstream tasks~\cite{melamud-etal-2016-context2vec,mccann2017learned,Peters2018DeepCW,Liu2019RoBERTaAR,Yang2019XLNetGA,Goldberg2019AssessingBS}.
It is also relatively easy to apply such pretrained LMs to comprehension tasks that are modeled as text span selection problems, where the boundary of an answer span can be predicted using a simple classifier on top of the  LM~\cite{Joshi2019SpanBERTIP}.


However, it is less clear how one could pretrain and fine-tune such models for other QA tasks that involve joint reasoning over both free-form NL text and \emph{structured} data.
One example task is semantic parsing for access to databases (DBs)~\citep{DBLP:conf/aaai/ZelleM96,berant2013freebase,DBLP:conf/acl/YihCHG15}, the task of transducing an NL utterance (\eg, \textit{``Which country has the largest GDP?''}) into a structured query over DB tables (\eg, SQL querying a database of economics).
A key challenge in this scenario is understanding the structured schema of DB tables (\eg, the name, data type, and stored values of columns), and more importantly, the alignment between the input text and the schema (\eg, the token \textit{``GDP''} refers to the {\tt Gross Domestic Product} column), which is essential for inferring the correct DB query~\citep{DBLP:dblp_conf/acl/BerantL14}.

Neural semantic parsers tailored to this task therefore attempt to learn joint representations of NL utterances and the (semi-)structured schema of DB tables (\eg, representations of its columns or cell values, as in~\citet{krishnamurthy17constraint,Bogin2019RepresentingSS,Wang2019RATSQLRS}, \textit{inter alia}).
However, this unique setting poses several challenges in applying pretrained LMs.
First, information stored in DB tables exhibit strong underlying structure, while existing LMs (\eg, BERT) are solely trained for encoding free-form text.
Second, a DB table could potentially have a large number of rows, and naively encoding all of them using a resource-heavy LM is computationally intractable.
Finally, unlike most text-based QA tasks (\eg, SQuAD, \newcite{Rajpurkar2016SQuAD10}) which could be formulated as a generic answer span selection problem and solved by a pretrained model with additional classification layers,
semantic parsing is highly domain-specific, and the architecture of a neural parser is strongly coupled with the structure of its underlying DB 
(\eg, systems for SQL-based and other types of DBs use different encoder models). 
In fact, existing systems have attempted to leverage BERT, but each with their own domain-specific, in-house strategies to encode the structured information in the DB~\citep{Guo2019TowardsCT,Zhang2019EditingBasedSQ,Hwang2019ACE},
and importantly, without pretraining representations on structured data.
These challenges call for development of 
general-purpose pretraining approaches tailored to learning representations for both NL utterances and structured DB tables. 

In this paper we 
present \model/, a pretraining approach for joint understanding of NL text and (semi-)structured tabular data (\autoref{sec:model}).
\model/ is built on top of BERT,
and jointly learns contextual representations for utterances and the structured schema of DB tables (\eg, a vector for each utterance token and table column).
Specifically, \model/ linearizes the structure of tables to be compatible with a Transformer-based BERT model. 
To cope with large tables, we propose \emph{content snapshots}, a method to encode a subset of table content most relevant to the input utterance.
This strategy is further combined with a \emph{vertical attention} mechanism to share information among cell representations in different rows (\autoref{sec:model:modeling}). 
To capture the association between tabular data and related NL text, \model/ is pretrained on a parallel corpus of 26 million tables and English paragraphs (\autoref{sec:model:pretraining}).


\model/ can be plugged into a neural semantic parser as a general-purpose encoder to compute representations for utterances and tables.
Our key insight is that although semantic parsers are highly domain-specific, most systems rely on representations of input utterances and the table schemas to facilitate subsequent generation of DB queries, and these representations can be provided by \model/, regardless of the domain of the parsing task.

We apply \model/ to two different semantic parsing paradigms: 
(1) a classical supervised learning setting on the \spider/ text-to-SQL dataset~\citep{Yu2018SpiderAL}, where \model/ is fine-tuned together with a task-specific parser using parallel NL utterances and labeled DB queries (\autoref{sec:exp:parser:spider});
and (2) a challenging weakly-supervised learning benchmark \wtq/~\citep{pasupat2015compositional}, where 
a system has to infer latent DB queries from its execution results (\autoref{sec:exp:parser:wtq}).
We demonstrate \model/ is effective in both scenarios, showing that it is a drop-in replacement of a parser's original encoder for computing contextual representations of NL utterances and DB tables.
Specifically, systems augmented with \model/ outperforms their counterparts using \bert/, registering state-of-the-art performance on \wtq/, while performing competitively on \spider/ (\autoref{sec:exp}).

\section{Background}


\paragraph{Semantic Parsing over Tables}
Semantic parsing tackles the task of translating an NL utterance $\utt$ into a formal meaning representation (MR) $\mr$.
Specifically, we focus on parsing utterances to access database tables, where $\mr$ is a structured query (\eg, an SQL query) executable on a set of relational DB tables $\mathcal{T}=\{ \tbl_t \}$.
A relational table $\tbl$ is a listing of $N$ rows $\{ \row_i \}_{i=1}^N$ of data, with each row $\row_i$ consisting of $M$ cells $\{ \cell_{\langle i, j \rangle} \}_{j=1}^{M}$, one for each column $\col_j$.
Each cell $\cell_{\langle i, j \rangle}$ contains a list of tokens. 

Depending on the underlying data representation schema used by the DB, a table could either be fully structured with strongly-typed and normalized contents (\eg, a table column named {\tt distance} has a unit of {\tt kilometers}, with all of its cell values, like \textit{200}, 
bearing the same unit), as is commonly the case for SQL-based DBs (\autoref{sec:exp:parser:spider}).
Alternatively, it could be semi-structured with unnormalized, textual cell values (\eg, \textit{200 km}, \autoref{sec:exp:parser:wtq}).
The query language could also take a variety of forms, from general-purpose DB access languages like SQL to domain-specific ones tailored to a particular task.

Given an utterance and its associated tables, a neural semantic parser generates a DB query from the vector representations of the utterance tokens and the structured schema of tables.
In this paper we refer \emph{schema} as the set of columns in a table, and its \emph{representation} as the list of vectors that represent its columns\footnote{Column representations for more complex schemas, \eg, those capturing inter-table dependency via primary and foreign keys, could be derived from these table-wise representations.}.
We will introduce how \model/ computes these representations in~\autoref{sec:model:modeling}. 

\begin{figure*}[t!]
	\centering
	\includegraphics[width=\textwidth]{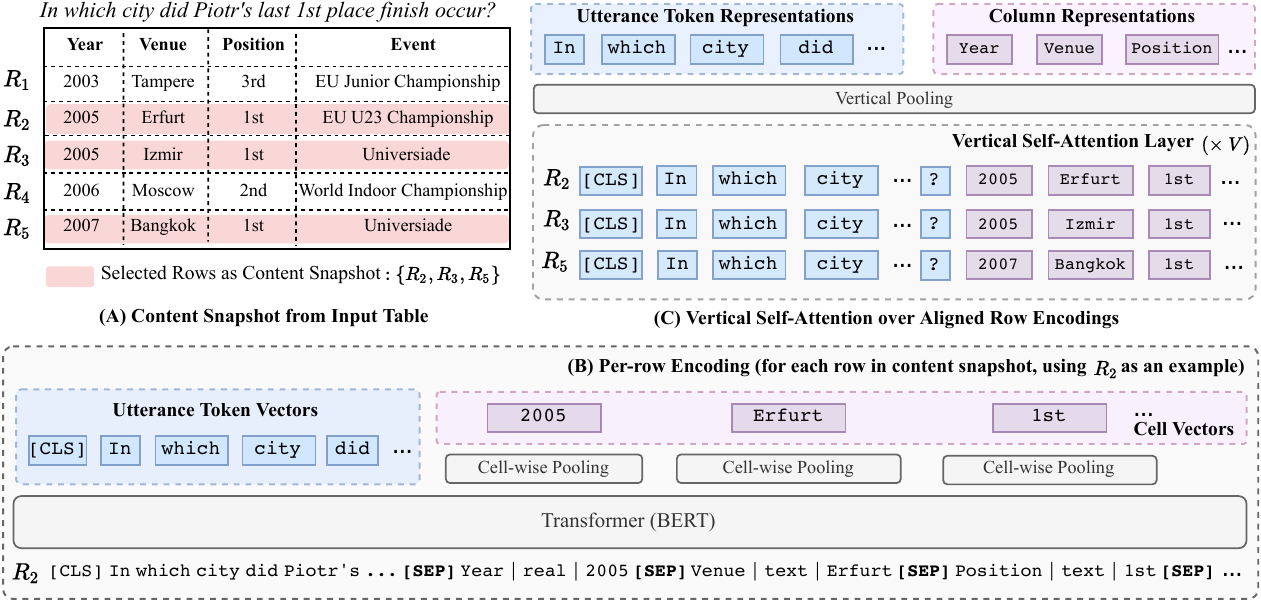}
	\caption{Overview of \model/ for learning representations of utterances and table schemas with an example from \wtq/\footnotemark. \textbf{(A)} A content snapshot of the table is created based on the input NL utterance. \textbf{(B)} Each row in the snapshot is encoded by a Transformer (only $R_2$ is shown), producing row-wise encodings for utterance tokens and cells. \textbf{(C)} All row-wise encodings are aligned and processed by $V$ vertical self-attention layers, generating utterance and column representations.}
	\label{fig:model:overview}
	\vspace{-3mm}
\end{figure*}


\paragraph{Masked Language Models}
Given a sequence of NL tokens $\xseq = x_1 , x_2, \ldots, x_n$, a masked language model (\eg, BERT) is an LM trained using the masked language modeling objective, which aims to recover the original tokens in $\xseq$ from a ``corrupted'' context created by randomly masking out certain tokens in $\xseq$. 
Specifically, let $\xseq_{m} = \{ x_{i_{1}}, \ldots, x_{i_m} \}$ be the subset of tokens
in $\xseq$ selected to be masked out, and \mask{\xseq} denote the masked sequence with tokens in $\xseq_m$ 
replaced by a {\tt [MASK]} symbol. A masked LM defines a  distribution $p_{\bm{\theta}}( \xseq_m | \mask{\xseq} )$ over the target tokens $\xseq_m$ given the masked context \mask{\xseq}.

BERT parameterizes $p_{\bm{\theta}}( \xseq_m | \mask{\xseq} )$ using a Transformer model.
During the pretraining phase, BERT maximizes $p_{\bm{\theta}}( \xseq_m | \mask{\xseq} )$ on large-scale textual corpora. 
In the fine-tuning phase, the pretrained model is used as an encoder to compute representations of input NL tokens, and its parameters are jointly tuned with other task-specific neural components.


\section{\model/: Learning Joint Representa- tions over Textual and Tabular Data}
\label{sec:model}

We first present how \model/ computes representations for NL utterances and table schemas (\autoref{sec:model:modeling}), and then describe the pretraining procedure (\autoref{sec:model:pretraining}).

\subsection{Computing Representations for NL Utterances and Table Schemas}
\label{sec:model:modeling}

\autoref{fig:model:overview} presents a schematic overview of \model/\footnotetext{Example adapted from \href{https://stanford.io/38iZ8Pf}{\tt stanford.io/38iZ8Pf}}.
Given an utterance $\utt$ and a table \tbl, \model/ first creates a \emph{content snapshot} of $\tbl$.
This snapshot consists of sampled rows that summarize the information in $\tbl$ most relevant to the input utterance.
The model then linearizes each row in the snapshot, concatenates each linearized row with the utterance, and uses the concatenated string as input to a Transformer (\eg, BERT) model, which outputs row-wise encoding vectors of utterance tokens and cells.
The encodings for all the rows in the snapshot are fed into a series of vertical 
self-attention layers, where a cell representation (or an utterance token representation) is computed by attending to vertically-aligned vectors of the same column (or the same NL token).
Finally, representations for each utterance token and column are generated from a pooling layer.

\paragraph{Content Snapshot}
One major feature of \model/ is its use of the table \emph{contents}, as opposed to just using the column names, in encoding the table schema.
This is motivated by the fact that contents provide more detail about the semantics of a column than just the column's name, which might be ambiguous.
For instance, the {\tt Venue} column in \autoref{fig:model:overview} which is used to answer the example question actually refers to \textit{host cities}, and encoding the sampled cell values while creating 
its representation may help 
match the term ``\textit{city}'' in the input utterance to this column.

However, a DB table could potentially have a large number of rows, with only few of them actually relevant to answering the input utterance.
Encoding all of the contents using a resource-heavy Transformer is both computationally intractable and likely not necessary.
Thus, we instead use a \emph{content snapshot} consisting of only a few rows that are most relevant to the input utterance, providing an efficient approach to calculate content-sensitive column representations from cell values.

We use a simple strategy to create content snapshots of $K$ rows based on the relevance between the utterance and a row.
For $K > 1$, we select the top-$K$ rows in the input table that have the highest $n$-gram overlap ratio with the utterance.\footnote{We use $n \leq 3$ in our experiments. Empirically this simple matching heuristic is able to correctly identify the best-matched rows for 40 out of 50 sampled examples on \wtq/.} 
For $K = 1$, to include in the snapshot as much information relevant to the utterance as possible, we create a synthetic row by selecting the cell values from each column that have the highest $n$-gram overlap with the utterance.
Using synthetic rows in this restricted setting is motivated by the fact that cell values most relevant to answer the utterance could come from multiple rows.
As an example, consider the utterance \textit{``How many more participants were there in \underline{2008} than in the \underline{London} Olympics?''}, and an associating table with columns {\tt Year}, {\tt Host City} and {\tt Number of Participants}, 
the most relevant cells to the utterance, {\tt 2008} (from {\tt Year}) and {\tt London} (from {\tt Host~City}), are from different rows, which could be included in a single synthetic row.
In the initial experiments we found synthetic rows also help stabilize learning.


\paragraph{Row Linearization}
\model/ creates a linearized sequence for each row in the content snapshot as input to the Transformer model.
\autoref{fig:model:overview}(B) depicts the linearization for $R_2$, which consists of a concatenation of the utterance, columns, and their cell values.
Specifically, each cell is represented by the name and data type\footnote{We use two data types, {\tt text}, and {\tt real} for numbers, predicted by majority voting over the NER labels of cell tokens.} of the column, together with its actual value, separated by a vertical bar. 
As an example, the cell $\cell_{\langle 2, 1 \rangle}$ valued {\tt 2005} in $\row_2$ in~\autoref{fig:model:overview} is encoded as
\begin{equation}\setlength\abovedisplayskip{4pt}\setlength\belowdisplayskip{4pt}
\small
	\underbrace{\text{\tt{Year}}}_{\textrm{Column Name}} \text{\tt|} \underbrace{\text{\tt{real}}}_{\textrm{Column Type}} \text{\tt|} ~~~ \underbrace{\text{\tt{2005}}}_{\textrm{Cell Value}}
	\label{eq:cell_string_representation}
\end{equation}
The linearization of a row is then formed by concatenating the above string encodings of all the cells, separated by the {\tt [SEP]} symbol. 
We then prefix the row linearization with utterance tokens as input sequence to the Transformer.

Existing works have applied different linearization strategies to encode tables with Transformers~\cite{Hwang2019ACE,Chen2019TabFactAL}, while our row approach is specifically designed for encoding content snapshots.
We present in~\autoref{sec:exp} results with different linearization choices.

\paragraph{Vertical Self-Attention Mechanism} 
The base Transformer model in \model/ outputs vector encodings of utterance and cell tokens for each row.
These row-level vectors are computed separately and therefore independent of each other.
To allow for information flow across cell representations of different rows, we propose vertical self-attention, a self-attention mechanism that operates over vertically aligned vectors from different rows.

As in~\autoref{fig:model:overview}(C), \model/ has $V$ stacked vertical-level self-attention layers.
To generate aligned inputs for vertical attention, we first compute a fixed-length initial vector for each cell at position $\langle i, j \rangle$, which is given by mean-pooling over the sequence of the Transformer's output vectors that correspond to its variable-length linearization as in Eq.~\eqref{eq:cell_string_representation}.
Next, the sequence of word vectors for the NL utterance (from the base Transformer model) are concatenated with the cell vectors as initial inputs to the vertical attention layer.

Each vertical attention layer has the same parameterization as the Transformer layer in~\cite{Vaswani2017AttentionIA}, but operates on vertically aligned elements, \ie, utterance 
and cell vectors 
that correspond to the same question token and column, respectively. 
This vertical self-attention mechanism enables the model to aggregate information from different rows in the content snapshot, allowing \model/ to capture cross-row dependencies on cell values.

\paragraph{Utterance and Column Representations}
A representation $\colvec_j$ is computed for each column $\col_j$ by mean-pooling over its vertically aligned cell vectors, $\{ \cellvec_{\langle i, j \rangle}: R_i \textrm{ in content snapshot} \}$, from the last vertical layer.
A representation for each utterance token, $\xvec_j$, is computed similarly over the vertically aligned token vectors.
These 
representations will be used by downstream neural semantic parsers.
\model/ also outputs an optional fixed-length table representation $\tblvec$ using the representation of the prefixed {\tt [CLS]} symbol, which is useful for parsers that operate on multiple DB tables.

\subsection{Pretraining Procedure}
\label{sec:model:pretraining}


\paragraph{Training Data} Since there is no large-scale, high-quality parallel corpus of NL text and structured tables, we instead use semi-structured tables that commonly exist on the Web as a surrogate data source.
As a first step in this line, we focus on collecting parallel data in English, while extending to multilingual scenarios would be an interesting avenue for future work. 
Specifically, we collect tables and their surrounding NL text from English Wikipedia and the WDC WebTable Corpus~\cite{Lehmberg2016ALP}, a large-scale table collection from CommonCrawl.
The raw data is extremely noisy, and we apply aggressive cleaning heuristics to filter out invalid examples (\eg, examples with HTML snippets or in foreign languages, and non-relational tables without headers).
See Appendix~\autoref{sec:app:pretrain:data} for details of data pre-processing.
The pre-processed corpus contains 26.6 million parallel examples of tables and NL sentences.
We perform sub-tokenization using the Wordpiece tokenizer shipped with BERT.




\paragraph{Unsupervised Learning Objectives} 
We apply different objectives for learning representations of the NL context and structured tables.
For NL contexts, we use the standard Masked Language Modeling (MLM) objective~\cite{Devlin2019BERTPO}, with a masking rate of 15\% sub-tokens in an NL context.

For learning column representations, we design two objectives motivated by the intuition that a column representation should contain both the general information of the column (\eg, its name and data type), and representative cell values relevant to the NL context.
First, 
a \textbf{Masked Column Prediction (MCP)} objective encourages the model to recover the names and data types of masked columns.
Specifically, we randomly select 20\% of the columns in an input table, masking their names and data types in each row linearization~(\eg, if the column {\tt Year} in \autoref{fig:model:overview} is selected, the tokens {\tt Year} and {\tt real} in Eq.~\eqref{eq:cell_string_representation} will be masked).
Given the column representation $\colvec_j$, \model/ is trained to predict the bag of masked (name and type) tokens from $\colvec_j$ using a multi-label classification objective.
Intuitively, MCP encourages the model to recover column information from its contexts.

Next, we use an auxiliary \textbf{Cell Value Recovery (CVR)} objective
to ensure information of representative cell values in content snapshots is retained after additional layers of vertical self-attention. 
Specifically, for each masked column $\col_j$ in the above MCP objective, CVR predicts the original tokens of each cell $\cell_{\langle i, j \rangle}$ (of $\col_j$) 
in the content snapshot
conditioned on its cell vector $\cellvec_{\langle i, j \rangle}$.\footnote{The cell value tokens are not masked in the input sequence, since predicting masked cell values is challenging even with the presence of its surrounding context. 
}
For instance, for the example cell $\cell_{\langle 2,1 \rangle}$ in Eq.~\eqref{eq:cell_string_representation}, we predict its value {\tt 2005} from $\cellvec_{\langle 2,1 \rangle}$.
Since a cell could have multiple value tokens, we apply the span-based prediction objective~\cite{Joshi2019SpanBERTIP}. 
Specifically, to predict a cell token $\cell_{{\langle i, j \rangle}_k} \in \cell_{{\langle i, j \rangle}}$, its positional embedding $\mathbf{e}_k$ and the cell representations $\cellvec_{\langle i, j \rangle}$ are fed into a two-layer network $f(\cdot)$ with GeLU activations~\cite{Hendrycks2016GaussianEL}.
The output of $f(\cdot)$ is then used to predict the original value token $\cell_{{\langle i, j \rangle}_k}$ from a softmax layer.


\section{Example Application: Semantic Parsing over Tables}
\label{sec:task}

We apply \model/ for representation learning on two semantic parsing paradigms, a classical supervised text-to-SQL task over structured DBs (\autoref{sec:exp:parser:spider}), and a weakly supervised parsing problem on semi-structured Web tables (\autoref{sec:exp:parser:wtq}).

\subsection{Supervised Semantic Parsing}
\label{sec:exp:parser:spider}

\paragraph{Benchmark Dataset}
Supervised learning is the typical scenario of learning a parser using parallel data of utterances and  queries.
We use \spider/~\cite{Yu2018SpiderAL}, a text-to-SQL dataset with 10,181 examples across 200 DBs.
Each example consists of an utterance~(\eg, \textit{``What is the total number of languages used in Aruba?''}), a DB with one or more tables, and an annotated SQL query, which typically involves joining multiple tables to get the answer (\eg, {\tt SELECT COUNT(*) FROM Country JOIN Lang ON Country.Code = Lang.CountryCode WHERE Name = `Aruba'}).

\paragraph{Base Semantic Parser}
We aim to show \model/ could help improve upon an already strong parser.
Unfortunately, at the time of writing, none of the top systems on \spider/ were publicly available.
To establish a reasonable testbed, we developed our in-house system based on TranX~\cite{yin18tranx}, an open-source general-purpose semantic parser.
TranX translates an NL utterance into an intermediate meaning representation guided by a user-defined grammar.
The generated intermediate MR could then be deterministically converted to domain-specific query languages (\eg, SQL). 

We use \model/ as encoder of utterances and table schemas.
Specifically, for a given utterance $\utt$ and a DB with a set of tables $\mathcal{T} = \{ \tbl_t \}$, we first pair $\utt$ with each table $\tbl_t$ in $\mathcal{T}$ as inputs to \model/, which generates $|\mathcal{T}|$ sets of table-specific representations of utterances and columns.
At each time step, an LSTM decoder performs hierarchical
attention~\cite{Libovick2017AttentionSF} over the list of table-specific representations, constructing an MR based on the predefined grammar.
Following the IRNet model~\cite{Guo2019TowardsCT} which achieved the best performance on \spider/, we use SemQL, a simplified version of the SQL, as the underlying grammar. 
We refer interested readers to Appendix~\autoref{sec:app:parser:spider} for details of our system. 

\subsection{Weakly Supervised Semantic Parsing}
\label{sec:exp:parser:wtq}

\paragraph{Benchmark Dataset}
Weakly supervised semantic parsing considers the reinforcement learning task of inferring the correct query from its execution results (\ie, whether the answer is correct).
Compared to supervised learning, weakly supervised parsing is significantly more challenging, as the parser does not have access to the labeled query, and has to explore the exponentially large search space of possible queries guided by the noisy binary reward signal of execution results.

\wtq/~\cite{pasupat2015compositional} is a popular dataset for weakly supervised semantic parsing, which has 22,033 utterances and 2,108 semi-structured Web tables from Wikipedia.\footnote{While some of the 421 testing Wikipedia tables might be included in our pretraining corpora, they only account for a very tiny fraction. In our pilot study, we also found pretraining only on Wikipedia tables resulted in worse performance.}
Compared to \spider/, examples in this dataset do not involve joining multiple tables, but typically require compositional, multi-hop reasoning over a series of entries in the given table (\eg, to answer the example in~\autoref{fig:model:overview} the parser needs to reason over the row set $\{ R_2, R_3, R_5 \}$, locating the {\tt Venue} field with the largest value of {\tt Year}).

\paragraph{Base Semantic Parser} MAPO~\cite{liang18mapo} is a strong system for weakly supervised semantic parsing.  
It improves the sample efficiency of the REINFORCE algorithm by biasing the exploration of queries towards the high-rewarding ones already discovered by the model.
MAPO uses a domain-specific query language tailored to answering compositional questions on single tables, and its utterances and column representations are derived from an LSTM encoder,
which we replaced with our \model/ model.
See Appendix~\autoref{sec:app:parser:wtq} for details of MAPO and our adaptation.

\section{Experiments}
\label{sec:exp}

In this section we evaluate \model/ on downstream tasks of semantic parsing to DB tables. 



\paragraph{Pretraining Configuration} 
We train two variants of the model, \modelbase/ and \modellarge/, with the underlying Transformer model initialized with the uncased versions of \bertbase/ and \bertlarge/, respectively.\footnote{We also attempted to train \model/ on our collected corpus from scratch without initialization from BERT, but with inferior results, potentially due to the average lower quality of web-scraped tables compared to purely textual corpora. We leave improving the quality of training data as future work.}
During pretraining, for each table and its associated NL context in the corpus, we create a series of training instances of paired NL sentences (as synthetically generated utterances) and tables (as content snapshots) by 
(1) sliding a (non-overlapping) context window of sentences with a maximum length of $128$ tokens, and 
(2) using the NL tokens in the window as the utterance, and pairing it with randomly sampled rows from the table as content snapshots. 
\model/ is implemented in PyTorch using distributed training.
Refer to Appendix \autoref{sec:app:pretrain:setup} for details of pretraining.


\paragraph{Comparing Models} 
We mainly present results for two variants of \model/ by varying the size of content snapshots $K$. 
\textbf{\model/$\mathbf{(K=3)}$} uses three rows from input tables as content snapshots and three vertical self-attention layers.
\textbf{\model/$\mathbf{(K=1)}$} uses one synthetically generated row as the content snapshot as described in \autoref{sec:model:modeling}. 
Since this model does not have multi-row input, we do not use additional vertical attention layers (and the cell value recovery learning objective). 
Its column representation $\colvec_j$ is defined by mean-pooling over the Transformer's output encodings that correspond to the column name (\eg, the representation for the {\tt Year} column in~\autoref{fig:model:overview} is derived from the vector of the {\tt Year} token in Eq.~\eqref{eq:cell_string_representation}).
We find this strategy gives better results compared with using the cell representation $\cellvec_j$ as $\colvec_j$.
%
We also compare with \textbf{\bert/} using the same row linearization and content snapshot approach as \model/\CS{1}, which reduces to a \model/\CS{1} model without pretraining on tabular corpora. 

\paragraph{Evaluation Metrics} As standard,
we report execution accuracy on \wtq/ and exact-match accuracy of DB queries on \spider/.

\subsection{Main Results}

\begin{table}[t]
\centering
\hspace{-3mm}
\resizebox{1.04 \columnwidth}{!}{%
\begin{tabular}{@{}lcccc@{}}
\toprule
\multicolumn{5}{c}{\textit{Previous Systems on WikiTableQuestions}} \\
Model  & \multicolumn{2}{c}{\textsc{Dev}} & \multicolumn{2}{c}{\textsc{Test}} \\ \hdashline
\citet{pasupat2015compositional}  & \multicolumn{2}{c}{37.0} & \multicolumn{2}{c}{37.1} \\
\citet{DBLP:journals/corr/NeelakantanLS15} & \multicolumn{2}{c}{34.1} & \multicolumn{2}{c}{34.2} \\
~~~~~~Ensemble 15 Models & \multicolumn{2}{c}{37.5} & \multicolumn{2}{c}{37.7} \\
\citet{Zhang2017MacroGA}  & \multicolumn{2}{c}{40.6} & \multicolumn{2}{c}{43.7} \\
\citet{Dasigi2019IterativeSF} & \multicolumn{2}{c}{43.1} & \multicolumn{2}{c}{44.3} \\
\citet{Agarwal2019LearningTG} & \multicolumn{2}{c}{43.2} & \multicolumn{2}{c}{44.1} \\ 
~~~~~~Ensemble 10 Models& \multicolumn{2}{c}{--} & \multicolumn{2}{c}{46.9} \\ 
\citet{wang19wikitable} & \multicolumn{2}{c}{43.7} & \multicolumn{2}{c}{44.5} \\
\hline
\multicolumn{5}{c}{\textit{Our System based on MAPO \citep{liang18mapo}}} \\
& \textsc{Dev} & Best & \textsc{Test} &  Best \\ 
Base Parser$^\dagger$  & 42.3 \std{0.3} & 42.7 & 43.1 \std{0.5} & 43.8 \\ \hdashline
~$w/$ \bertbase/ \CS{1} & 49.6 \std{0.5} & 50.4 & 49.4 \std{0.5} & 49.2 \\
~~~~~~ $-$ content snapshot & 49.1 \std{0.6} & 50.0 & 48.8 \std{0.9} & 50.2 \\
~$w/$ \modelbase/ \CS{1} & 51.2 \std{0.5} & 51.6 & 50.4 \std{0.5} & 51.2 \\
~~~~~~ $-$ content snapshot & 49.9 \std{0.4} & 50.3 & 49.4 \std{0.4} & 50.0 \\
~$w/$ \modelbase/ \CS{3} & 51.6 \std{0.5} & 52.4 & 51.4 \std{0.3} & 51.3 \\ \hdashline

~$w/$ \bertlarge/ \CS{1} & 50.3 \std{0.4} & 50.8 & 49.6 \std{0.5} & 50.1 \\
~$w/$ \modellarge/ \CS{1} & 51.6 \std{1.1} & 52.7 & 51.2 \std{0.9} & 51.5 \\
~$w/$ \modellarge/ \CS{3} & \textbf{52.2} \std{0.7} & \textbf{53.0} & \textbf{51.8} \std{0.6} & \textbf{52.3} \\
\bottomrule
\end{tabular}}
\caption{Execution accuracies on \wtq/. $^\dagger$Results from \citet{liang18mapo}. (\textsc{Ta})\textsc{Bert} models are evaluated with 10 random runs. We report mean, standard deviation and the best results. \textsc{Test$\mapsto$Best} refers to the result from the run with the best performance on \textsc{Dev.}~set.}
\label{tab:results:wtq}
\vspace{-3mm}
\end{table}

\begin{table}[t]
\centering
\resizebox{0.97 \columnwidth}{!}{%
\begin{tabular}{lcc}
\toprule
\multicolumn{3}{@{}c@{}}{
	\begin{tabular}{@{}lc@{}}
		\multicolumn{2}{c}{\textit{Top-ranked Systems on Spider Leaderboard}} \\ 
		Model & \textsc{Dev.~Acc.} \\ \hdashline
        Global--GNN~\citep{Bogin2019GlobalRO} & 52.7 \\
        EditSQL $+$ \bert/~\citep{Zhang2019EditingBasedSQ} & 57.6 \\
        RatSQL~\citep{Wang2019RATSQLRS}  & 60.9  \\ 
        IRNet $+$ \bert/~\citep{Guo2019TowardsCT}  & 60.3  \\
        ~~$+$ Memory $+$ Coarse-to-Fine  & 61.9  \\ 
        IRNet V2 $+$ \bert/  & 63.9  \\ 
        RyanSQL $+$ \bert/~\citep{choi20ryansql}  & \textbf{66.6}  \\
	\end{tabular}
} \\

\hline
\multicolumn{3}{c}{\textit{Our System based on TranX \citep{yin18tranx}}} \\
 & Mean & Best \\  \hdashline
$w/$ \bertbase/ \CS{1} & 61.8 \std{0.8} & 62.4 \\
~~~~~~ $-$ content snapshot & 59.6 \std{0.7} & 60.3 \\
$w/$ \modelbase/ \CS{1} & 63.3 \std{0.6} & 64.2 \\
~~~~~~ $-$ content snapshot & 60.4 \std{1.3} & 61.8 \\
$w/$ \modelbase/ \CS{3} & 63.3 \std{0.7} & 64.1 \\ \hdashline

$w/$ \bertlarge/ \CS{1} & 61.3 \std{1.2} & 62.9 \\
$w/$ \modellarge/ \CS{1} & 64.0 \std{0.4} & 64.4 \\
$w/$ \modellarge/ \CS{3} & \textbf{64.5} \std{0.6} & \textbf{65.2} \\
\bottomrule
\end{tabular}}
\caption{Exact match accuracies on the public development set of \spider/. Models are evaluated with 5 random runs. }
\label{tab:results:spider}
\vspace{-3mm}
\end{table}

\autoref{tab:results:wtq} and \autoref{tab:results:spider} summarize the end-to-end evaluation results on \wtq/ and \spider/, respectively.
First, comparing with existing strong semantic parsing systems, we found our parsers with \model/ as the utterance and table encoder perform competitively.
On the test set of \wtq/, MAPO augmented with a \modellarge/ model with three-row content snapshots, \modellarge/\CS{3}, registers a single-model exact-match accuracy of 52.3\%, surpassing the previously best ensemble system (46.9\%) from \citet{Agarwal2019LearningTG} by 5.4\% absolute.

On \spider/, our semantic parser based on TranX and SemQL (\autoref{sec:exp:parser:spider}) is conceptually similar to the base version of IRNet as both systems use the SemQL grammar, 
while our system has a simpler decoder.
Interestingly, we observe that its performance with \bertbase/ (61.8\%) matches the full BERT-augmented IRNet model with a stronger decoder using augmented memory and coarse-to-fine decoding (61.9\%).
This confirms that our base parser is an effective baseline.
Augmented with representations produced by \modellarge/\CS{3}, our parser 
achieves up to 65.2\% exact-match accuracy, a 2.8\% increase over the base model using \bertbase/.
Note that while other competitive systems on the leaderboard use BERT with more sophisticated semantic parsing models, our best \textsc{Dev.}~result is already close to the score registered by the best submission (RyanSQL$+$\bert/).
This suggests that if they instead used \model/ as the representation layer, they would see further gains.

Comparing semantic parsers augmented with \model/ and \bert/, we found \model/ is more effective across the board.
We hypothesize that the performance improvements would be attributed by two factors. 
First, pre-training on large parallel textual and tabular corpora helps \model/ learn to encode structure-rich tabular inputs in their linearized form (Eq.~\eqref{eq:cell_string_representation}), whose format is different from the ordinary natural language data that \bert/ is trained on.
Second, pre-training on parallel data could also helps the model produce representations that better capture the alignment between an utterance 
and the relevant information presented in the structured schema,
which is important for semantic parsing.

Overall, the results on the two benchmarks demonstrate that pretraining on aligned 
textual and tabular data is necessary for joint understanding of NL utterances and tables,
and \model/ works well with both structured (\spider/) and semi-structured (\wtq/) DBs, and agnostic of the task-specific structures of semantic parsers.



\paragraph{Effect of Content Snapshots} 
In this paper we propose using content snapshots to capture the information in input DB tables that is most relevant to answering the NL utterance.
We therefore study the effectiveness of including content snapshots when generating schema representations. 
We include in \autoref{tab:results:wtq} and \autoref{tab:results:spider} results of models without using content in row linearization (``$-$content snapshot'').
Under this setting a column is represented as ``{\tt Column Name | Type}'' without cell values (\textit{c.f.}, Eq.~\eqref{eq:cell_string_representation}).
We find that content snapshots are helpful for both \bert/ and \model/, especially for \model/.
As discussed in~\autoref{sec:model:modeling}, encoding sampled values from columns in learning their representations helps the model infer alignments between entity and relational phrases in the utterance and the corresponding  column.
This is particularly helpful for identifying relevant columns from a DB table that is mentioned in the input utterance.
As an example, empirically we observe that on \spider/ our semantic parser with \modelbase/ using just one row of content snapshots \CS{1} registers a higher accuracy of selecting the correct columns when generating SQL queries (\eg, columns in {\tt SELECT} and {\tt WHERE} clauses), compared to the \modelbase/ model without encoding content information (87.4\% v.s.~86.4\%).

\begin{table}[t]
    \centering
    \resizebox{\columnwidth}{!}{%
    \begin{tabular}{@{}l|l|l|l@{}}
        \toprule
        \multicolumn{4}{@{}c@{}}{$\utt$: \textit{How many years before was the film \textcolor{amaranth}{\underline{Bacchae}} out before \textcolor{amaranth}{\underline{the Watermelon}?}}}  \\ \midrule
        \multicolumn{4}{@{}c@{}}{Input to \modellarge/ \CS{3} \hfill $\triangleright$ \textit{Content Snapshot with Three Rows }} \\
        \textbf{Film} & \textbf{Year} & \textbf{Function} & \textbf{Notes} \\ \hdashline
        \textcolor{amaranth}{\underline{The Bacchae}} & 2002 & Producer & Screen adaptation of... \\
        The Trojan Women & 2004 & Producer/Actress & Documutary film... \\
        \textcolor{amaranth}{\underline{The Watermelon}} & 2008 & Producer & Oddball romantic comedy... \\ \midrule
        \multicolumn{4}{@{}c@{}}{Input to \modellarge/ \CS{1} \hfill $\triangleright$ \textit{Content Snapshot with One Synthetic Row }}\\
        \textbf{Film} & \textbf{Year} & \textbf{Function} & \textbf{Notes} \\ \hdashline
        \textcolor{amaranth}{\underline{The Watermelon}} & 2013 & Producer & Screen adaptation of... \\ \bottomrule
    \end{tabular}}
    \caption{Content snapshots generated by two models for a \wtq/ \textsc{Dev.}~example. Matched tokens between the question and content snapshots are \textcolor{amaranth}{\underline{underlined}}. }
    \label{tab:exp:content_snapshot:example}
    \vspace{-3mm}
\end{table}

Additionally, comparing \model/ using one synthetic row \CS{1} and three rows from input tables \CS{3} as content snapshots, the latter generally performs better.
Intuitively, encoding more table contents relevant to the input utterance could potentially help answer questions that involve reasoning over information across multiple rows in the table.
\autoref{tab:exp:content_snapshot:example} shows such an example, and to answer this question a parser need to subtract the values of {\tt Year} in the rows for \textit{``The Watermelon''} and \textit{``The Bacchae''}. 
\modellarge/ \CS{3} is able to capture the two target rows in its content snapshot and generates the correct DB query, while the \modellarge/\CS{1} model with only one row as content snapshot fails to answer this example. 

\paragraph{Effect of Row Linearization}  
\model/ uses row linearization to represent a table row as sequential input to Transformer. 
\autoref{tab:exp:linearization} (upper half) presents results using various linearization methods.
We find adding type information and content snapshots 
improves performance, as they provide more hints about the meaning of a column.

We also compare with existing linearization methods in literature using a \modelbase/ model, with results shown in \autoref{tab:exp:linearization} (lower half). 
\citet{Hwang2019ACE} uses BERT to encode concatenated column names to learn column representations.
In line with our previous discussion on the effectiveness content snapshots,
this simple strategy without encoding cell contents underperforms 
(although with \modelbase/ pretrained on our tabular corpus the results become slightly better).
Additionally, we remark that linearizing table contents has also be applied to other BERT-based tabular reasoning tasks.
For instance, \citet{Chen2019TabFactAL} propose a ``natural'' linearization approach for checking if an NL statement entails the factual information listed in a table using a binary classifier with representations from \bert/, where a table is linearized by concatenating the semicolon-separated cell linearization for all rows. Each cell is represented by a phrase ``{\tt \underline{column name} is \underline{cell value}}''.
For completeness, we also tested this cell linearization approach,
and find \bertbase/ achieved improved results.
We leave pretraining \model/ with this linearization strategy as promising future work.


\begin{table}[t]
    \centering
    \resizebox{\columnwidth}{!}{%
    \begin{tabular}{lcc}
    \toprule
    Cell Linearization Template & \textsc{WikiQ.} & \textsc{Spider} \\ \hline
      \multicolumn{3}{c}{Pretrained \modelbase/ Models \CS{1}} \\
      {\tt \underline{Column Name}} & 49.6 \std{0.4} & 60.0 \std{1.1}  \\
      {\tt \underline{Column Name} | \underline{Type}}$^\dagger$ \hfill ($-$content snap.) & 49.9 \std{0.4} & 60.4 \std{1.3} \\
      {\tt \underline{Column Name} | \underline{Type} | \underline{Cell Value}}$^\dagger$ & 51.2 \std{0.5} & 63.3 \std{0.6} \\ \hdashline
      \multicolumn{3}{c}{ \bertbase/ Models} \\
      {\tt \underline{Column Name}} \cite{Hwang2019ACE} & 49.0 \std{0.4} & 58.6 \std{0.3} \\
      {\tt \underline{Column Name} is \underline{Cell Value}}~\hyperlink{TabFact}{(Chen19)} & 50.2 \std{0.4} & 63.1 \std{0.7} \\ \bottomrule
    \end{tabular}}
    \caption{Performance of pretrained \modelbase/ models and \bertbase/ on the \textsc{Dev.} sets with different linearization methods. \underline{Slot names} are underlined. $^\dagger$Results copied from~\autoref{tab:results:wtq} and \autoref{tab:results:spider}.} 
    \label{tab:exp:linearization}
    \vspace{-3mm}
\end{table}

\begin{table}[t]
    \centering
    \resizebox{0.75 \columnwidth}{!}{%
    \begin{tabular}{lcc}
    \toprule
    Learning Objective & \textsc{WikiQ.} & \textsc{Spider} \\ \hline
      MCP only &  51.6 \std{0.7}  & 62.6 \std{0.7} \\ 
      MCP + CVR & 51.6 \std{0.5} & 63.3 \std{0.7} \\ \bottomrule
    \end{tabular}}
    \caption{Performance of pretrained \modelbase/\CS{3} on \textsc{Dev.}~sets with different pretraining objectives.} 
    \label{tab:exp:objective}
    \vspace{-3mm}
\end{table}

\paragraph{Impact of Pretraining Objectives} 
\model/ uses two objectives (\autoref{sec:model:pretraining}), a masked column prediction (MCP) and a cell value recovery (CVR) objective, to learn column representations that could capture both the general information of the column (via MCP) and its representative cell values related to the utterance (via CVR).
\autoref{tab:exp:objective} shows ablation results of pretraining \model/ with different objectives.
We find \model/ trained with both MCP and the auxiliary CVR objectives gets a slight advantage, suggesting CVR could potentially lead to more representative column representations with additional cell information.

\section{Related Works}

\paragraph{Semantic Parsing over Tables}
Tables are important media of world knowledge.
Semantic parsers have been adapted to operate over structured DB  tables~\cite{wang15overnight,xu2017sqlnet,dong18coarsefine,Yu2018SyntaxSQLNetST,Shi2018IncSQLTI,Wang2018RobustTG}, and open-domain, semi-structured Web tables~\cite{pasupat2015compositional,Sun2016TableCS,DBLP:journals/corr/NeelakantanLS15}.
To improve representations of utterances and tables for neural semantic parsing, existing systems have applied pretrained word embeddings (\eg., GloVe, as in \citet{DBLP:journals/corr/abs-1709-00103,yu18typesql,sun18sql,liang18mapo}), and BERT-family models for learning joint contextual representations of utterances and tables, but with domain-specific approaches to encode the structured information in tables~\cite{Hwang2019ACE,He2019XSQLRS,Guo2019TowardsCT,Zhang2019EditingBasedSQ}.
\model/ advances this line of research by presenting a general-purpose, pretrained encoder over parallel corpora of Web tables and NL context.
Another relevant direction is to augment representations of columns from an individual table with global information of its linked tables defined by the DB schema~\cite{Bogin2019GlobalRO,Wang2019RATSQLRS}.
\model/ could also potentially improve performance of these systems with improved table-level representations.

\paragraph{Knowledge-enhanced Pretraining} 
Recent pre-training models have incorporated structured information from knowledge bases (KBs) or other structured semantic annotations 
into training contextual word representations, either by fusing vector representations of entities and relations on KBs into word representations of LMs~\cite{Peters2019KnowledgeEC,Zhang2019ERNIEEL,Zhang2019SemanticsawareBF}, or by encouraging the LM to recover KB entities and relations from text~\cite{Sun2019ERNIEER,Liu2019KBERTEL}.
\model/ is broadly relevant to this line in that it also exposes an LM with structured data (\ie, tables), while aiming to learn joint representations for both textual and structured tabular data.

\section{Conclusion and Future Work}
We present \model/, a pretrained encoder for joint understanding of textual and tabular data.
We show that semantic parsers using \model/ as a general-purpose feature representation layer achieved strong results on two benchmarks.
This work also opens up several avenues for future work.
First, we plan to evaluate \model/ on other related tasks involving joint reasoning over textual and tabular data (\eg, table retrieval and table-to-text generation).
Second, following the discussions in~\autoref{sec:exp}, we will explore other table linearization strategies with Transformers, improving the quality of pretraining corpora, as well as novel unsupervised objectives.
Finally, to extend \model/ to cross-lingual settings with utterances in foreign languages and structured schemas defined in English, we plan to apply more advanced semantic similarity metrics for creating content snapshots.


\bibliography{tablebert}
\bibliographystyle{acl_natbib}

\newpage
\clearpage
\appendix
\onecolumn
\begin{center}
\Large
\model/: Pretraining for Joint Understanding of Textual and Tabular Data

Supplementary Materials
\end{center}

\section{Pretraining Details}

\subsection{Training Data}
\label{sec:app:pretrain:data}

We collect parallel examples of tables and their surrounding NL sentences from two sources:
\paragraph{Wikipedia Tables} We extract all the tables 
on English Wikipedia\footnote{We do not use infoboxes (tables on the top-right of a Wiki page that describe properties of the main topic), as they are not relational tables.}. For each table, we use the preceding three paragraphs as the NL context, as we observe that most Wiki tables are located after where they are described in the body text.

\paragraph{WDC WebTable Corpus}~\cite{Lehmberg2016ALP} is a large collection of Web tables extracted from the Common Crawl Web scrape\footnote{\href{http://webdatacommons.org/webtables}{\tt http://webdatacommons.org/webtables}}. We use its 2015 English-language relational subset, which consists of $50.8$ million relational tables and their surrounding NL contexts. 

\paragraph{Preprocessing}
Our dataset is collected from arbitrary Web tables, which are extremely noisy.
We develop a set of heuristics to clean the data by: 
(1) removing columns whose names have more than 10 tokens; 
(2) filtering cells with more than two non-ASCII characters or 20 tokens;
(3) removing empty or repetitive rows and columns; 
(4) filtering tables with less than three rows and four columns, and 
(5) running {\tt spaCy} to identify the data type of columns (text or real value) by majority voting over the NER labels of column tokens, 
(6) rotating vertically oriented tables.
We sub-tokenize the corpus using the Wordpiece tokenizer in~\citet{Devlin2019BERTPO}.
The pre-processing results in 1.3 million tables from Wikipedia and 25.3 million tables from the WDC corpus.

\subsection{Pretraining Setup}
\label{sec:app:pretrain:setup}
As discussed in \autoref{sec:exp}, we create training instances of NL sentences (as synthetic utterances) and content snapshots from tables by sampling from the parallel corpus of NL contexts and tables.
Each epoch contains 37.6$M$ training instances. 
We train \model/ for 10 epochs.
\autoref{tab:app:hyperparameters} lists the hyper-parameters used in training.
Learning rates are validated on the development set of \wtq/.
We use a batch size of 512 for large models to reduce training time.
The training objective is sum of the three pretraining objectives in \autoref{sec:model:pretraining} (Masked Language Modeling objective for utterance tokens, Masked Column Prediction\footnote{An exception is that for pretraining \model/\CS{1} models, the masked column prediction objective reduces to the vanilla masked language modeling objective since there are no additional vertical attention layers.} and Column Value Recovery objectives for columns and their cell values).
Our largest model \modellarge/\CS{3} takes six days to train for 10 epochs on 128 Tesla V100 GPUs using mixed precision training.

\begin{table}[H]
	\centering
	\resizebox{0.8 \columnwidth}{!}{%
	\begin{tabular}{lcccc}
	\toprule
	\textbf{Parameter} & \modelbase/\CS{1} & \modellarge/\CS{1} & \modelbase/\CS{3} &  \modellarge/\CS{3} \\
	\midrule
	Batch Size & 256 & 512 & 512 & 512 \\
	Learning Rate & $2 \times 10^{-5}$ & $2 \times 10 ^{-5}$ & $4 \times 10^{-5}$ & $4 \times 10^{-5}$ \\
	Max Epoch & \multicolumn{4}{c}{10} \\
	Weight Decay & \multicolumn{4}{c}{0.01} \\
	Gradient Norm Clipping & \multicolumn{4}{c}{1.0} \\
	\bottomrule
	\end{tabular}}
	\caption{Hyper-parameters using in pretraining}
	\label{tab:app:hyperparameters}
\end{table}

\section{Semantic Parsers}
\label{sec:app:parser}

\subsection{Supervised Parsing on \spider/}
\label{sec:app:parser:spider} 

\paragraph{Model} We develop our text-to-SQL parser based on TranX~\cite{yin18tranx}, which translates an NL utterance into a tree-structured abstract meaning representation following user-specified grammar, before deterministically convert the generated abstract MR into an SQL query. TranX models the construction process of an abstract MR (tree-structured representation of an SQL query) using a transition-based system, which decomposes its generation story into a sequence of actions following the user defined grammar.

Formally, given an input NL utterance $\utt$ and a database with a set of tables $\mathcal{T} = \{ \tbl_i \}$, the probability of generating of an SQL query (\ie, its semantically equivalent MR) 
$\mr$ is decomposed as the production of action probabilities:
\begin{equation}
  p(\mr|\utt, \mathcal{T}) = \prod p(a_t | a_{<t}, \utt, \mathcal{T})
\end{equation}
where $a_t$ is the action applied to the hypothesis at time stamp $t$. $a_{<t}$ denote the previous action history. 
We refer readers to~\citet{yin18tranx} for details of the transition system and how individual action probabilities are computed. 
In our adaptation of TranX to text-to-SQL parsing on \spider/, 
we follow~\citet{Guo2019TowardsCT} and use SemQL as the underlying grammar, which is a simplification of the SQL language. 
\autoref{fig:spider:grammar} lists the SemSQL grammar specified using the abstract syntax description language~\cite{wang97asdl}.
Intuitively, the generation starts from a tree-structured derivation with the root production rule {\tt select\_stmt}$\mapsto${\tt SelectStatement}, which lays out overall the structure of an SQL query.
At each time step, the decoder algorithm locates the current opening node on the derivation tree, following a depth-first, left-to-right order.
If the opening node is not a leaf node, the decoder invokes an action $a_t$ which expands the opening node using a production rule with appropriate type.
If the current opening node is a leaf node (\eg, a node denoting string literal),
the decoder fills in the leaf node using actions that emit terminal values.

\lstset{%
  basicstyle=\fontfamily{cmtt}\small,
  columns=fullflexible,
  frame=bt,
  morecomment=[l][\color{red}]{\#},
  escapechar=\&
  }
\newcommand*{\Comment}[1]{\hfill\makebox[8.0cm][l]{\textcolor{red}{\# #1}}}%

\begin{figure}[t]
    \centering
    \begin{lstlisting}
select_stmt = SelectStatement(
    distinct distinct,                  &\Comment{DISTINCT keyword}&
    expr* result_columns,               &\Comment{Columns in SELECT clause}&
    expr? where_clause,                 &\Comment{WHERE clause}&
    order_by_clause? order_by_clause,   &\Comment{ORDER BY clause}&
    int? limit_value,                   &\Comment{LIMIT clause}&
    table_ref* join_with_tables,        &\Comment{Tables in the JOIN clause}&
    compound_stmt? compound_statement   &\Comment{Compound statements (\eg, UNION, EXCEPT)}&
)

distinct = None | Distinct

order_by_clause = OrderByClause(expr* expr_list, order order)

order = ASC | DESC

expr = AndExpr(expr* expr_list)
      | OrExpr(expr* expr_list)
      | NotExpr(expr expr)
      | CompareExpr(compare_op op, expr left_value, expr right_value)
      | AggregateExpr(aggregate_op op, expr value, distinct distinct)
      | BinaryExpr(binary_op op, expr left_value, expr right_value)
      | BetweenExpr(expr field, expr left_value, expr right_value)
      | InExpr(column_ref left_value, expr right_value)
      | LikeExpr(column_ref left_value, expr right_value)
      | AllRows(table_ref table_name)
      | select_stmt
      | Literal(string value)
      | ColumnReference(column_ref column_name)

aggregate_op = Sum | Max | Min | Count | Avg

compare_op = LessThan | LessThanEqual | GreaterThan 
            | GreaterThanEqual | Equal | NotEqual

binary_op = Add | Sub | Divide | Multiply

compound_stmt = CompoundStatement(compound_op op, select_stmt query)

compound_op = Union | Intersect | Except
\end{lstlisting}
    \caption{ASDL Grammar of SemQL used in TranX}
    \label{fig:spider:grammar}
\end{figure}

To use such a transition system to generate SQL queries, we extend its action space with two new types of actions, \textsc{SelectTable}$(\tbl_i)$ for node of type {\tt table\_ref} in \autoref{fig:spider:grammar}, which selects a table $\tbl_i$ (\eg, for predicting target tables for a {\tt FROM} clause), and \textsc{SelectColumn}$(\tbl_i, c_j)$ for node of type {\tt column\_ref}, which selects the column $c_j$ from table $\tbl_i$ (\eg, for predicting a result column used in the {\tt SELECT} clause).

As described in~\autoref{sec:exp:parser:spider}, \model/ produces a list of entries, with one entry $\langle \tblvec_i, \mathbf{X}_i, \mathbf{C}_i \rangle$ for each table $\tbl_i$: 
\begin{equation}
	\mathbb{M} = \Big\{  \langle \tblvec_i, \mathbf{X}_i = \{ \xvec_1, \xvec_2, \ldots \}, \mathbf{C}_i = \{ \colvec_1, \colvec_2, \ldots,  \} \rangle_i  \Big\}_{i=1}^{|\mathcal{T}|}
\end{equation}
where each entry $\langle \tblvec_i,  \mathbf{X}_i, \mathbf{C}_i \rangle$ in $\mathbb{M}$ 
consists of $\tblvec_i$, the representation of table $\tbl_i$ given by the output vector of the prefixed {\tt [CLS]} symbol, the table-specific representations of utterance tokens $\mathbf{X}_i = \{ \xvec_1, \xvec_2, \ldots \}$, 
and representations of columns in $\tbl_i$, $\mathbf{C}_i =\{ \colvec_1, \colvec_2, \ldots \}$. 
At each time step $t$, the decoder in TranX performs hierarchical attention over representations in $\mathbb{M}$ to compute a context vector.
First, a table-wise attention score is computed using the LSTM's previous state, $\textbf{state}_{t-1}$ with the set of table representations.
\begin{equation}
	\mathrm{score}(\tbl_i) = \mathrm{Softmax}\Big(\mathrm{DotProduct}(\mathbf{state}_{t-1}, \mathrm{key}( \tblvec_i ) ) \Big),
	\label{eq:app:parser:select_table_action}
\end{equation}
where the linear projection $\mathrm{key}(\cdot) \in \mathbb{R}^{256}$ projects the table representations to key space. Next, for each table $\tbl_i \in \mathcal{T}$, a table-wise context vector $\mathbf{ctx}(\tbl_i)$ is generated by attending over the union of vectors in utterance token representations $\mathbf{X}_i$ and column representations $\mathbf{C}_i$:
\begin{equation}
	\mathbf{ctx}(\tbl_i) = \mathrm{DotProductAttention}\Big(\mathbf{state}_{t-1}, \mathrm{key}(\mathbf{X}_i \cup \mathbf{C}_i), \mathrm{value} ( \mathbf{X}_i \cup \mathbf{C}_i ) \Big),
\end{equation}
with the LSTM state as the query, $\mathrm{key}(\cdot)$ as the key, and another linear transformation $\mathrm{value} (\cdot) \in \mathbb{R}^{256}$ to project the representations to value vectors. 
The final context vector is then given by the weighted sum of these table-wise context vectors $\mathbf{ctx}(\tbl_i)$ ($i \in \{ 1, \ldots, |\mathcal{T}| \}$) weighted by the attention scores $\mathrm{score}(\tbl_i)$. The generated context vector is then used to update the state of the decoder LSTM to $\mathbf{state}_t$.

The updated decoder state is then used to compute the probability of carrying out the action defined at time step $t$, $a_t$.
For a \textsc{SelectTable}$(\tbl_i)$ action, its probability of is defined similarly as Eq.~\eqref{eq:app:parser:select_table_action}.
For a \textsc{SelectColumn}$(\tbl_i, c_j)$ action, it is factorized as the probability of selecting the table $\tbl_i$ (given by Eq.~\eqref{eq:app:parser:select_table_action}), times the probability of selecting the column $c_j$. The latter is defined as
\begin{equation}
	\mathrm{score}(c_j) = \mathrm{Softmax}\Big( \mathrm{DotProduct} ( \mathbf{state}_t, {\colvec_j} ) \Big).
\end{equation}

We also add simple entity linking features to the representations in $\mathbb{M}$, defined by the following heuristics: 
(1) If an utterance token $x \in \utt$ matches with the name of a table $\tbl$, we concatenate a trainable embedding vector ($\mathbf{table\_match} \in \mathbb{R}^{16}$) to the representations of $x$ and $\tbl$.
(2) Similarly, we concatenate an embedding vector ($\mathbf{column\_match} \in \mathbb{R}^{16}$) to the representations of an utterance token and a column if their names match.
(3) Finally, we concatenate a zero-vector ($\mathbf{0} \in \mathbb{R}^{16}$) to representations of all unmatched elements.

\paragraph{Configuration} We use the default configuration of TranX. For \model/ parameters, we use an Adam optimizer with a learning rate of $3e-5$ and linearly decayed learning rate schedule, and another Adam optimizer with a constant learning rate of $1e-3$ for all remaining parameters. 
During training, we update model parameters for 25000 iterations, and freeze the \model/ parameters at the first 1000 update steps. We use a batch size of 30 and beam size of 3. 
We use gradient accumulation for large models to fit a batch into GPU memory.

\subsection{Weakly-supervised Parsing on \wtq/}
\label{sec:app:parser:wtq}

\paragraph{Model} We use MAPO~\cite{liang18mapo}, a strong weakly-supervised semantic parser.
The original MAPO models comes with an LSTM encoder, which generates utterance and column representations used by the decoder to predict table queries.
We directly substitute the encoder with \model/, and project the utterance and table representations from \model/ to the original embedding space using a linear transformation.
MAPO uses a domain-specific query language tailored to answer compositional questions on a single table.
For instance, the example question in \autoref{fig:model:overview} could be answered using the following query
\begin{lstlisting}
Table.contains(column=Position, value=1st)  &\Comment{Get rows whose `Position' field contains `1st'}&  
      .argmax(order_by=Year)  &\Comment{Get the row which has the largest `Year' field}&
      .hop(column=Venue)     &\Comment{Select the value of `Venue' in the result row}&
\end{lstlisting}
MAPO is written in Tensorflow. In our experiments we use an optimized re-implementation in PyTorch, which yields 4$\times$ training speedup.

\paragraph{Configuration} We use the same optimizer and learning rate schedule as in~\autoref{sec:app:parser:spider}. We use a batch size of 10, and train the model for 20000 steps, with the \model/ parameters frozen at the first 5000 steps. Other hyper-parameters are kept the same as the original MAPO system.

\end{document}